\documentclass[]{spie}  
\usepackage{color}
\usepackage[]{graphicx}
\usepackage{xcolor}
\usepackage{hyperref}

\title{Extracting associations and meanings of objects depicted in artworks through bi-modal deep networks}

\author{
 Gregory Kell,\supit{a} Ryan-Rhys Griffiths,\supit{b} Anthony Bourached,\supit{c} \\ and David G. Stork\supit{d} \\
\supit{a} Department of Population Health,
King\rq s College London, London, UK \\
\supit{b} Department of Physics, University of Cambridge, Cambridge, UK\\
\supit{c} Oxia Palus, London, UK \\
\supit{d} Consultant, Portola Valley, CA 94028 USA
}

\begin{document}

\maketitle

\begin{abstract}
  We present a novel bi-modal system based on deep networks to address the problem of learning associations and simple meanings of objects depicted in \lq\lq authored\rq\rq\ images, such as fine art paintings and drawings.  Our overall system processes both the images and associated texts in order to learn associations between images of individual objects, their identities and the abstract meanings they signify.  Unlike past deep nets that {\em describe} depicted objects and infer predicates, our system identifies meaning-bearing objects (\lq\lq signifiers\rq\rq ) and their associations (\lq\lq signifieds\rq\rq ) as well as basic overall meanings for target artworks.  Our system had precision of $48\%$ and recall of $78\%$ with an F1 metric of $0.6$ on a curated set of Dutch {\em vanitas} paintings, a genre celebrated for its concentration on conveying a meaning of great import at the time of their execution.  We developed and tested our system on fine art paintings but our general methods can be applied to other authored images.
\end{abstract}

\section{Introduction:  Extracting meanings conveyed through images}

Despite recent successes in semantic image analysis,\cite{Andersonetal:18,Badrinarayananetal:17} a large class of challenging image analysis problems have yet to be addressed by automatic methods, specifically inferring {\em meanings} expressed in images.  While traditional semantic image analysis can learn and represent the identities of physical objects, their relations, and general context (indoor/outdoor, day/night), it has yet to address adequately the problem of inferring the motives and messages of an image\rq s creator or \lq\lq author.\rq\rq\ \ 

The general problem of extracting a plausible meaning from an arbitrary authored image is surely AI complete.  Specifically, such meanings are context dependent, and viewers bring a wealth of general commonsense knowledge and domain-specific information when interpreting such images.  This context information is generally represented in automatic systems somewhat abstractly and symbolically, ultimately derived from texts.  It is for such reasons that our system, described in Sect.~\ref{sec:Architecture}, relies on both natural language processing {\em and} image analysis;  it also relies on a bi-modal knowledge representation that facilitates simple inference based on both image and symbolic or textual knowledge.    

The overwhelming majority of natural photographs used in semantic image analysis record a scene but carry few if any specific messages beyond the obvious, such as ({\sc I like this} or {\sc I was here}).  By contrast, that vast majority of Western artworks were crafted to convey a message, moral, or {\em meaning}, often at patrons\rq\ requests.    For example, the Catholic Church, as patron, commissioned numerous artists to depict stories, episodes and messages from the Bible, to adorn religious sites and documents.  Such messages ranged from specific, such as {\sc Adam and Eve disobeyed an order and committed mankind\rq s original sin}, or more ones that are more general, such as {\sc have compassion for those less fortunate}.  Likewise, Hindu artworks convey morals such as {\sc without being attached to the fruits of activities, one should act as a matter of duty}; in Greek mythology {\sc Zeus controls your fate}; in political art {\sc communism is evil}; and in advertising {\sc buy Coca-cola}.  No photograph---natural or \lq\lq art\rq\rq\ photograph---has a complexity and depth of meaning as some masterpiece paintings, such as Diego Vel{\`a}zquez\rq s {\em Las meninas}, which is the most analysed painting in all of art.\cite{StantonPruitt:02}  

A key challenge to AI is that any one of these messages might be conveyed through innumerable images that at least on the surface bear little relation to one another.  One need merely think of all the images in a print advertising campaign for life insurance policies to appreciate such a breadth.  While there may be multiple plausible readings of a work, some readings are consistent with more information, or the \lq\lq ground truth\rq\rq\ of an author\rq s stated goals.  
Below we shall address this problem in a restricted domain:  the {\em vanitas} artworks of the Dutch Golden Age.  These works were created with clear intention to convey religious and moral messages.  Note, though, that we address this art genre because the problem domain is well circumscribed---in the objects that are depicted, the styles, and messages.  We stress the general problem is broader than just the analysis of this genre:  other artwork, commercial advertisements, political propaganda, and some information graphics convey meaning or import that may be amenable to our methods.

Our work presented here is inspired by recent work using deep networks that address a related problem:  identifying key figures or \lq\lq actors\rq\rq\ in paintings as a step to understanding a work\rq s meaning.\cite{Storketal:21a}  That previous approach automatically identified and located key signs or iconographic {\em attributes} in an artwork, and found the closest segmented human figure.  Thus, for instance, a segmented human figure would be automatically identified as St.~John the Baptist because of its proximity to his iconographic attribute, the crucifixion cross;  likewise Christ was identified by his proximity to one of his attributes, a dove;  and so on.  

Semantic analysis of images involves relating images to concepts expressed in text.  In that prior work, the representation and data structure linking images with text consisted of a list of associations of semiotic attributes with actors (specifically saints depicted in paintings)---a relation that was entered by hand.  In the system reported in the present paper, such an association was learned through natural language processing of texts associated with art images.

\section{Semiotics:  Signifiers and signfieds} \label{sec:VisualSemiotics}

The nature, representation, and reasoning related to meaning is central to the branch of philosophy known as epistemology, to the field of semiotics, and to much of cognitive science and artificial intelligence.  Most work in AI on inferring meanings has been restricted to documents and natural-language text, in large part because almost all texts were written by authors intending to convey information and non-trivial meaning, while the same cannot be said for the majority of natural photographers.\cite{Targon:18}  Computational approaches to visual semiotics are not as well developed as the corresponding approaches to text analysis, but has nevertheless shed light on a number of problems.\cite{Aiello:19}

The essential items of concern in semiology are {\em signifiers} and {\em signifieds}.  Meaning derives from relations and associations between certain objects, marks, or works---signifiers---and their associated objects, concepts, or ideas---signifieds.  There are three basic classes of signifiers:

\begin{description}
    \item[Signs] In principle, signs can be anything, so long as there is some conceptual link from them to a signified.  For instance a smashed wineglass on the floor may refer to the (absent) person who threw it there.
    \item[Icons] Icons resemble their associated signifieds but are simpler and more abstract.  For instance the digital pixel-based icons of printers and file folders on computer desktops refer to physical printers and folders and, indirectly, their function.
    \item[Symbols] Symbols are arbitrary conventions, often created by stakeholders as, for instance {\bf \$}, which indicates {\sc dollars}, or more generally {\sc money}.
\end{description}

All three classes of signifiers appear in some paintings and support the creation of meaning of such works.

\subsection{Herman van~Steenwijk\rq s {\em Still life:  An allegory of the vanities of human life}}

Consider an example of how signifiers and signifieds work in Herman van~Steenwijk\rq s {\em Still life:  An allegory of the vanities of human life}.  The most compelling message of the work is:  {\sc Do not concern yourself with life of this world, but instead prepare for eternal life to come}.  Note that all educated viewers (and most illiterate viewers) in 17th-century Netherlands would understand this message, which was also preached from pulpits, widely written and discussed.  Indeed, this message was associated with the Dutch Revolt and ultimate break from its rulers in Catholic Spain.  

\begin{figure} \label{fig:SteenwijkAllegory}
  \centering
  \includegraphics[width=.6\textwidth]{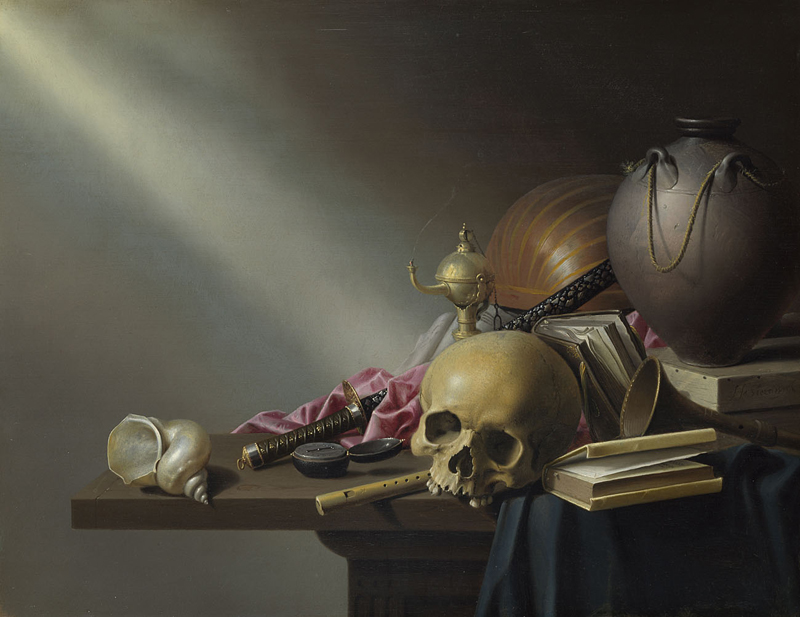}
  \caption{Herman van~Steenwijk\rq s {\em Still life:  An allegory of the vanities of human life} (c.~1640).  Nearly every object in this work is a sign and contributes to the overall message and meaning of the work, which is fairly explicit in the work\rq s subtitle.  Moreover its highly realistic style and composition (eg., with light shaft directed toward the skull) support the work\rq s primary meaning.}
\end{figure}

How is that message conveyed, and how might an automated system extract or infer such a message?

The skull is a sign that refers to mortality and the inevitability of death;  the books are signs that refer to worldly (not otherworldly) knowledge;  the musical instruments---flute, sackbut (early trombone) and lute (early guitar)---are signs that refer to culture and sensual pleasures;  the lovely seashell and samurai sword refer to worldly travel and luxury;  the open pocket watch refers to the passage of time and human life;  the oil lamp at the rear refers to the passage of time and life and the fact that it has recently been extinguished (as evident from the thin trail of smoke) refers to the fact that life can end at any moment.  Finally, the shaft of light directed toward the skull has religious connotations as well.

Our overall technical goal is to compute simple meanings given such a vanitas painting, specifically the references or significations of objects and relations withing the painting to simple meanings.  Our computational approach is to {\em learn} these signifier-signified relations from existing texts describing such artworks and use that information as part of a system that identifies these meanings of objects within the artworks.  Specifically, our system will learn relations such as {\sc skull:mortality}, {\sc book:worldly knowledge}, and so on, as can be represented in a knowledge graph, as shown in  Fig.~\ref{fig:KnowledgeGraph}.  Further, computer vision and pattern recognition will identify the signifiers from the image alone. 
To the best of our knowledge, the work we present here is the first to use natural language processing and computer vision to extract simple meanings from authored images.

\section{Description of task and models}

Our overall task is to develop a system that takes an art image as input and performs simple interpretation of its meaning, particularly the meaning of depicted objects.  Thus, for instance, such a system would allow a user to point to an object in a painting and ask \lq\lq What does that mean?\rq\rq\ and the system would not only provide the  identity of an object but also its simple meaning, for example {\sc that is a skull, which represents mortality}.

\subsection{System components}

In order to compute the below interpretation of the work in Fig.~\ref{fig:BernaertStillLife}, the model would need to not only detect the objects in the image, but also associate them to the relevant concepts.  For instance, the lute, music and inkstand would need to be recognised and associated with the creative endeavours by the same system. Thus, the system requires an object recognition component that detects the signifiers and a knowledge graph that associates the signifiers with the signifieds.  The knowledge graph can be automatically created by leveraging natural processing techniques to extract information from texts describing the symbolism of vanitas artworks and using the relevant entities and relations to create a knowledge graph.

The accompanying text describes the work in Fig.~\ref{fig:BernaertStillLife}: 

\begin{figure}
    \centering
    \includegraphics[scale=0.1]{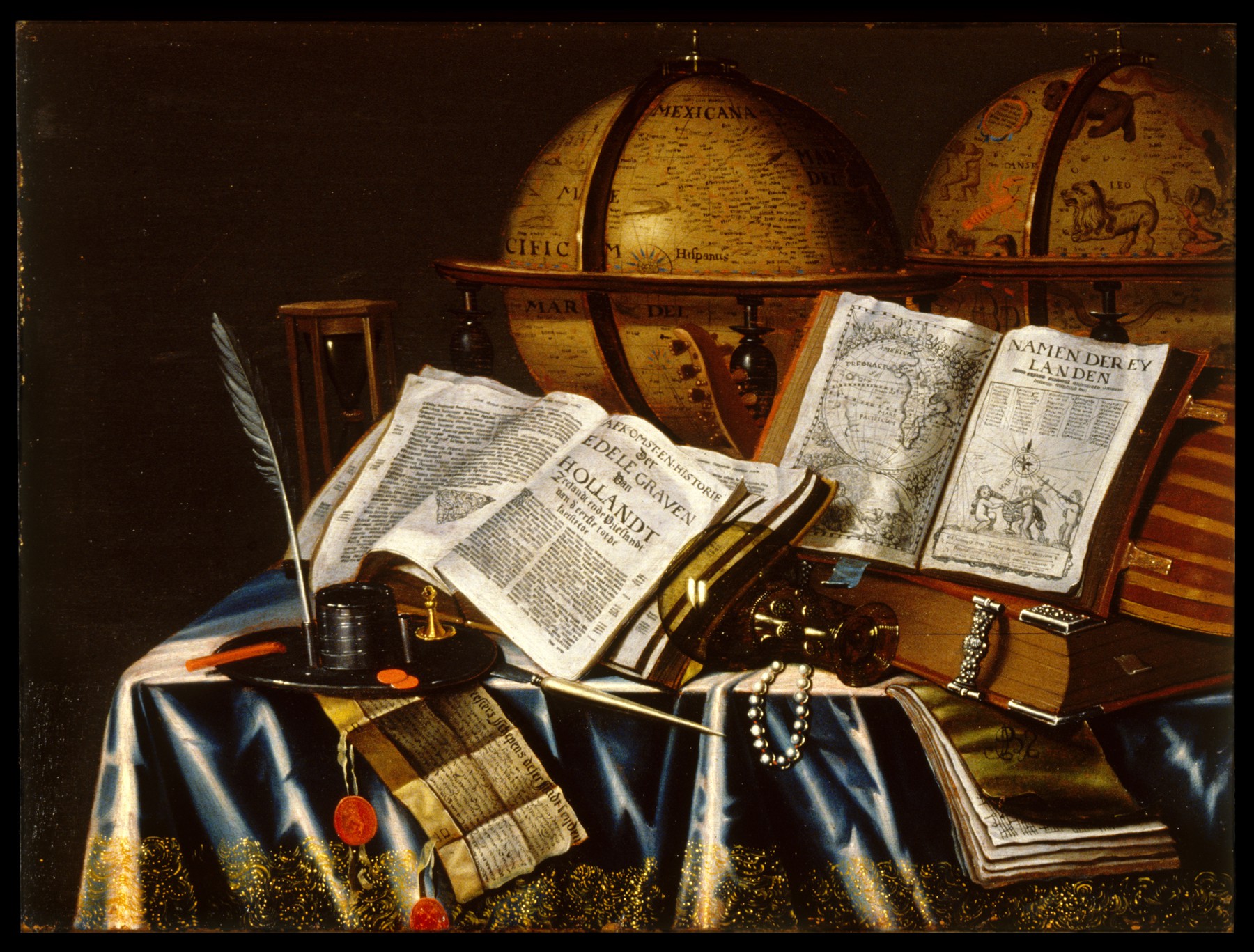}
    \caption{Adam Bernaert\rq s \textit{Still Life} (1665).  }
    \label{fig:BernaertStillLife}
\end{figure}

\begin{figure}
    \centering
    \includegraphics[scale=0.1]{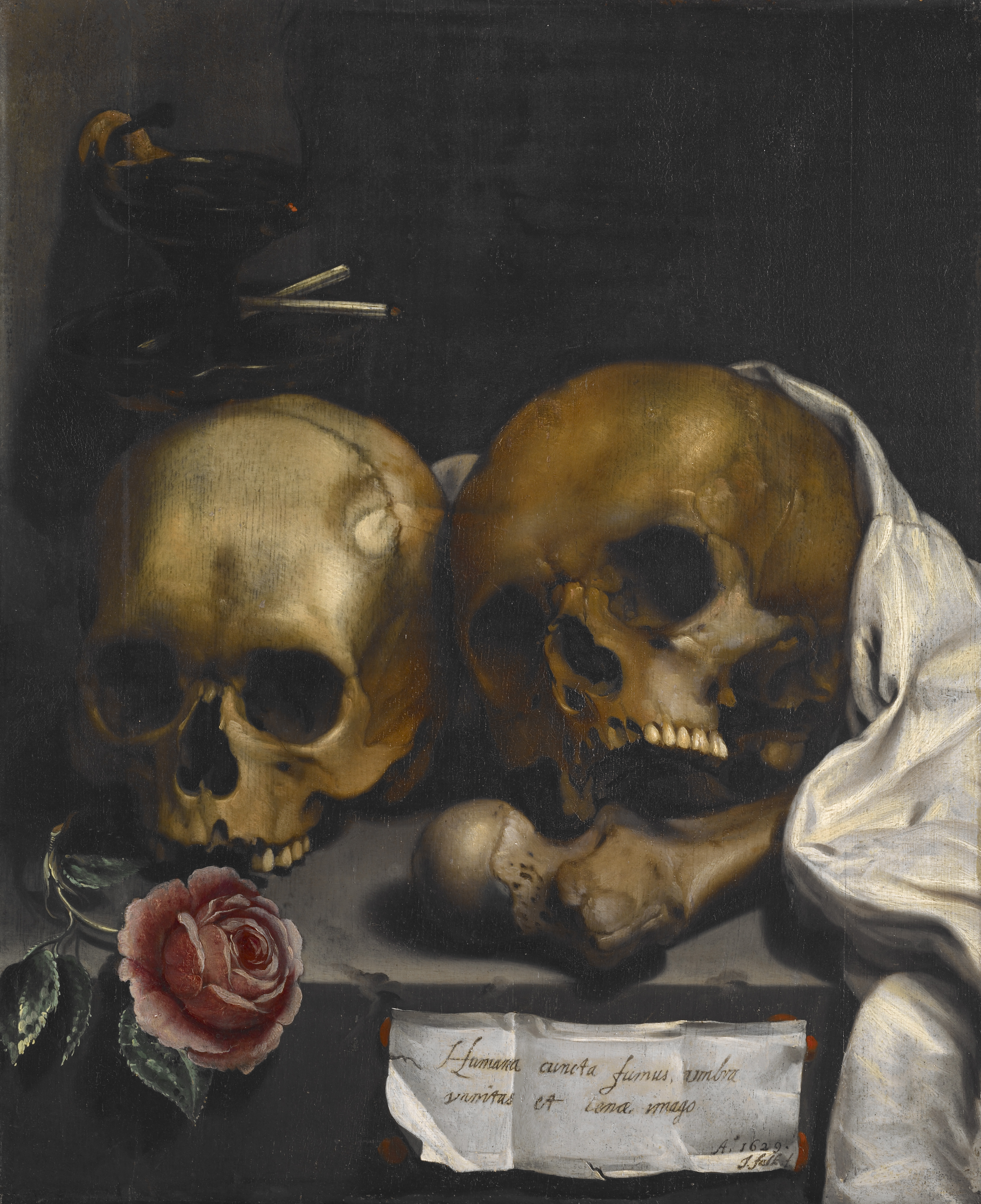}
    \caption{\textit{Vanitas Still Life
J Falk (Dutch, 1600-1699)} The symbolism of the skulls in this painting is obvious, but the rose (quick to wilt) and oil lamp (easily snuffed out) also refer to life’s brevity and fragility. The vanitas symbolism is underscored by the Latin inscription underneath: \lq\lq All that is human is smoke, show, vanity and the picture of a stage.\rq\rq }
    \label{fig:FalkVanitasStillLife}
\end{figure}

\begin{figure}
    \centering
    \includegraphics[scale=0.05]{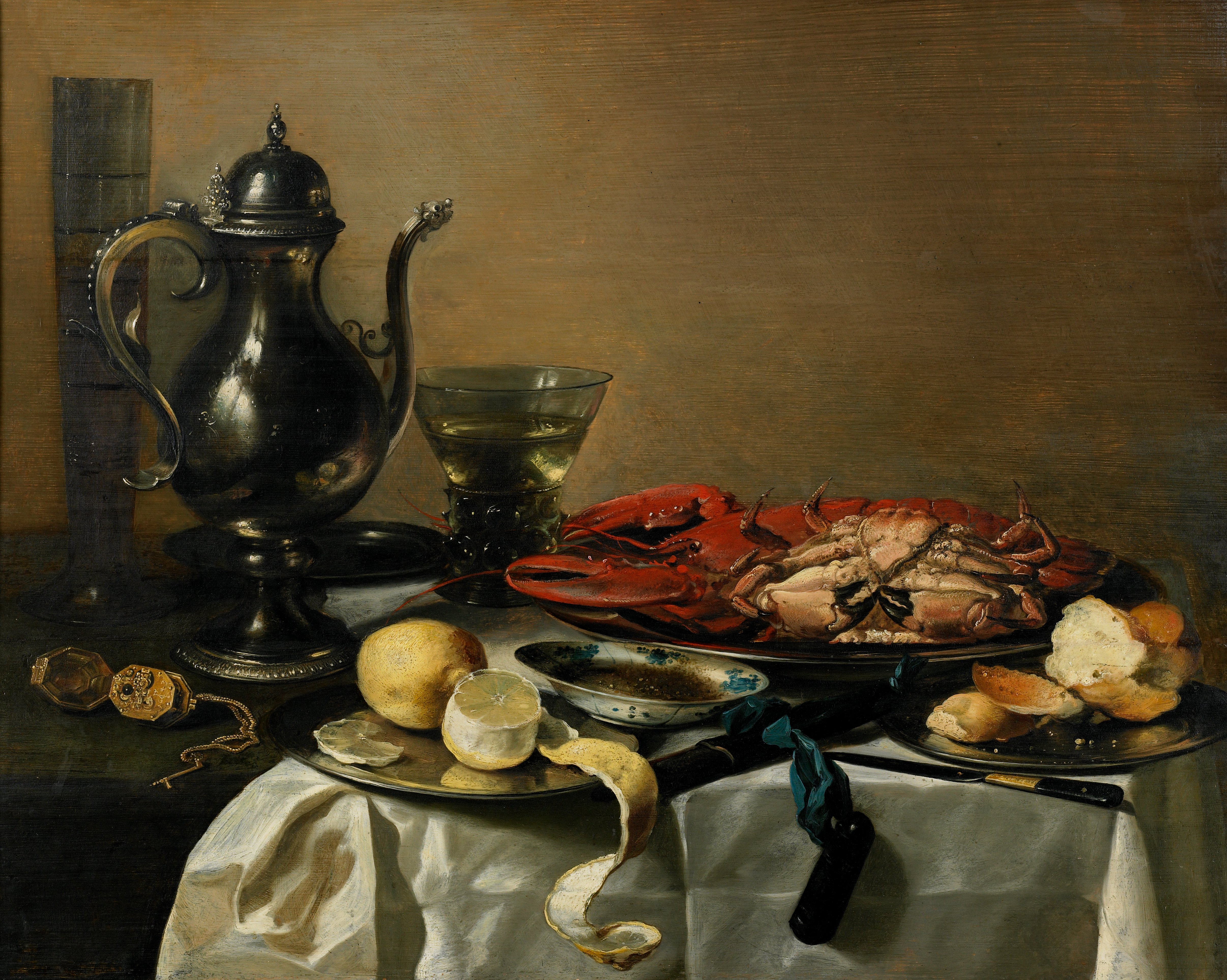}
    \caption{\textit{Still Life, 1643
Pieter Claesz.} Everything on the table, from the fluted glass and goblet to the lobster and crab, is indeed life-like. You can almost smell the lemons. The Dutch proudly displayed such expensive status symbols in their homes, the exotic food and material possessions reminding them of the good things in life, even as the watch reminds them of their transience. The bread and wine, in a touch of Christian symbolism, echo the moralizing message of vanitas, or vanity: all earthly things must pass.}
    \label{fig:ClaeszStillLife}
\end{figure}

\begin{figure}
    \centering
    \includegraphics[scale=0.3]{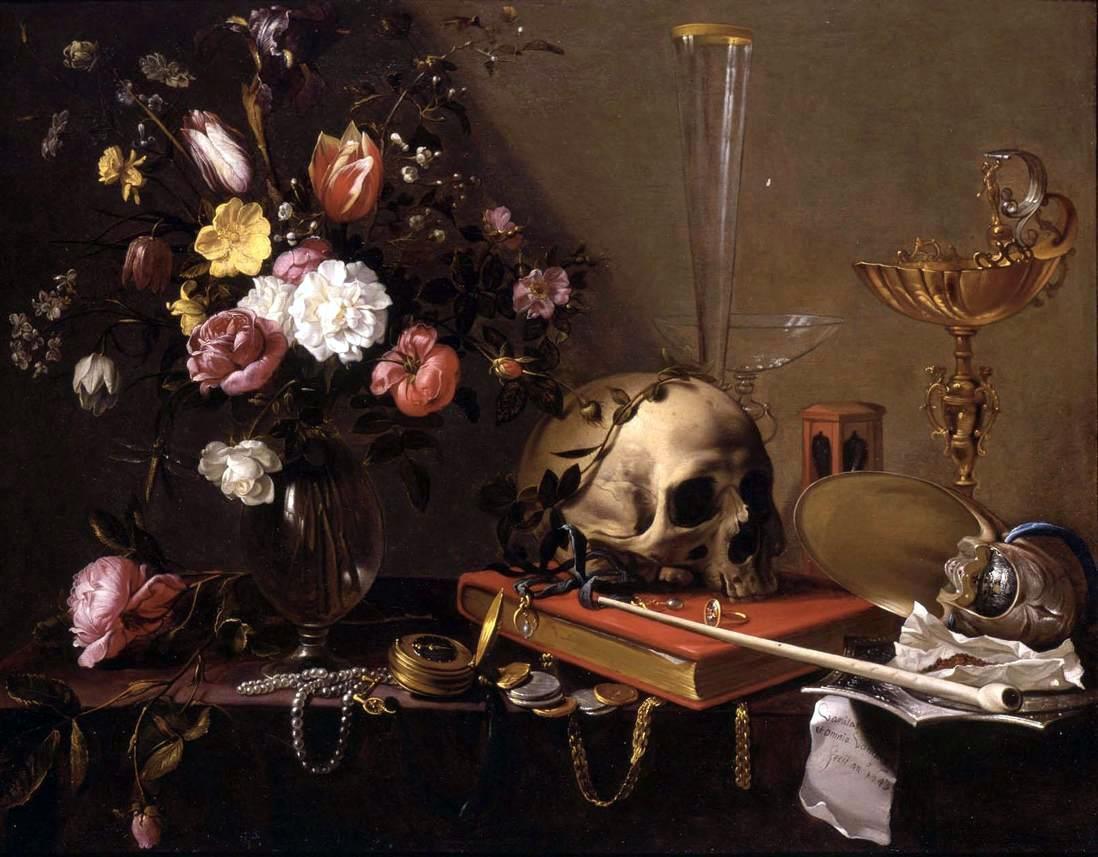}
    \caption{\textit{1642, Vanitas Still Life with Flowers and Skull, Adriaen van Utrecht} In his 1642 painting, Vanitas Still Life with Flowers and Skull, Adriaen van Utrecht depicts a multitude of objects, including but not limited to a vase of flowers, a human skull, small gold and silver coins, two glass vases, and a book. In the tradition of still-life painting, these objects have individual meanings all their own. For example, the orange book beneath the skull symbolizes human knowledge. Oftentimes in still-life paintings artists would choose to paint books with their pages open, exposing their content to the viewer as a window into the intellectual realm (note 1). In front of the book, draping softly over the edge of the table, are several gold and silver coins, representations of wealth and status. Another manifestation of the vanitas theme can be found in the timepiece that rests next to the scintillating currency. The cover of this gold clock remains open for the viewer to tell the time and be reminded, once again, about the inevitable passing of time. Although integral to the composition, these symbolic objects do not expose the intrinsic meaning of Utrecht’s still life. Indeed, this painting’s meaning transcends these objects’ symbolism, and it can arguably be found in relationship forged between the bouquet and the skull. \\ The light radiating from the bouquet of flowers on the left side of the canvas serves as a point of entry into Utrecht’s cluttered, frenetic composition. Of the flower varietals represented in the glass vase, the bright white and yellow flowers are first to captivate the viewer’s attention. Their vibrancy refuses to be ignored, and light seems to glow from the depths of their petals. The flowers found at the base of the glass vase seem to have withered away and fallen from the cohesive bouquet above. One of the rose buds, along with the metal chains and a scrap of paper, hangs limp over the table’s edge, boldly entering the viewer’s space and creating a trompe l’oeil effect. Most of the light in this painting falls on the bouquet and the skull, likely an intentional way for Utrecht to highlight the two most important objects of the painting. Prior to becoming prominent elements within still-life paintings, skulls were oftentimes painted on the back of portraits to remind their owners of life’s brevity (note 2). Placing a flower arrangement directly next to a human skull was surely intentional on the part of the artist, and their proximity forces a dialogue about the relationship between life and death. A craggy, decaying skull in the presence of vibrant, budding flowers seems unsettling at first, and initially the viewer cannot help but wonder how the two could ever be related. A closer look, however, reveals that a profound connection does, indeed, exist between the two ostensibly polar opposites. As the title indicates, Utrecht’s painting is replete with evidence of the theme of vanitas, a theme that tenderly reminds us about the transience of life. Skulls are particularly interesting in that they are no longer part of a living human, nor are they really an inanimate object (note 3). They fall somewhere in the middle, emphasizing their power as symbols of vanitas and reminders to cherish life before death. Utrecht’s work calls the viewer beyond vanitas, however. More than merely symbols of knowledge, life and death, these objects invite the viewer to contemplate the culture of cultivation in which they once existed.}
    \label{fig:UtrechtVanitasStillLife}
\end{figure}

\begin{figure}
    \centering
    \includegraphics[scale=0.1]{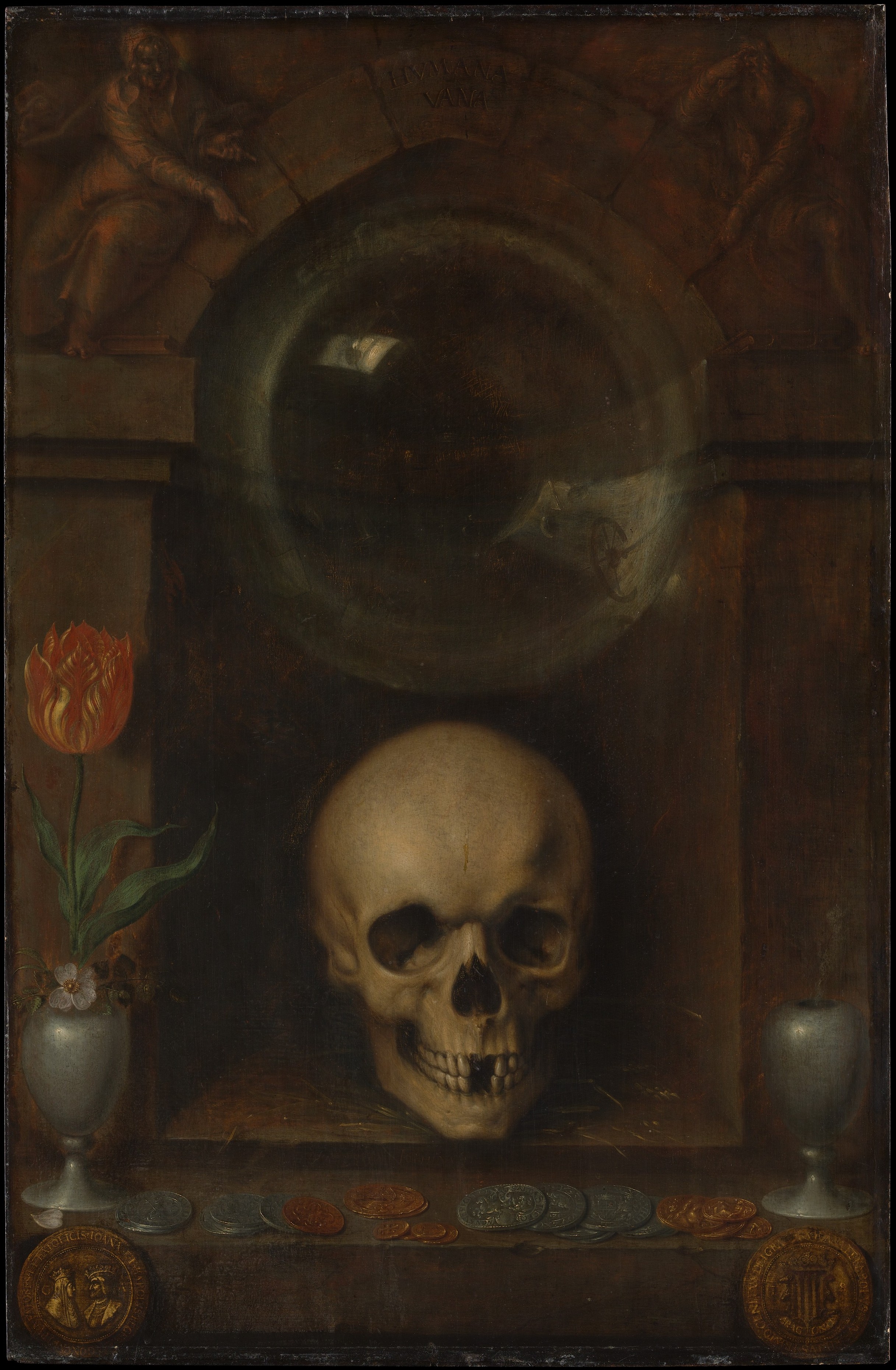}
    \caption{\textit{Vanitas Still Life, 1603, Jacques de Gheyn II } De Gheyn was a wealthy amateur who is best known as a brilliant draftsman, but he also painted and engraved. This panel is generally considered to be the earliest known independent still-life painting of a vanitas subject, or symbolic depiction of human vanity. The skull, large bubble, cut flowers, and smoking urn refer to the brevity of life, while images floating in the bubble—such as a wheel of torture and a leper’s rattle—refer to human folly. The figures flanking the arch above are Democritus and Heraclitus, the laughing and weeping philosophers of ancient Greece.}
    \label{fig:GheynStillLife}
\end{figure}

\begin{figure}
    \centering
    \includegraphics[scale=0.1]{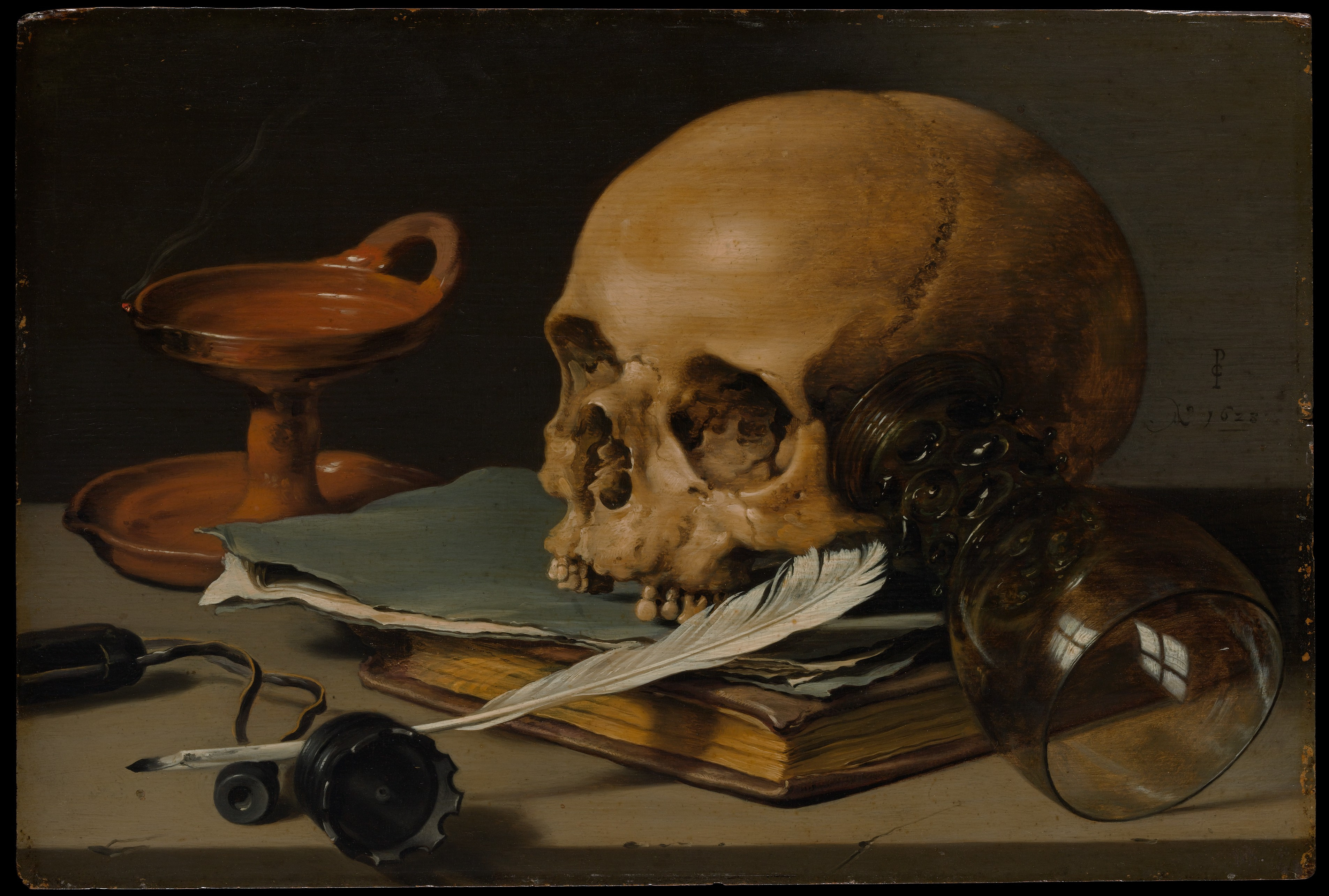}
    \caption{\textit{Still Life with a Skull and a Writing Quill, 1628, Pieter Claesz} In this still life, close observation and realistic detail operate in tension with explicit symbolism. The toppled glass, gap-toothed skull, and guttering wick of an oil lamp all serve as stark symbols of life’s brevity. Working with a limited palette of grays and browns, Claesz carefully describes the surfaces of these unsettling objects. By arranging them on a pitted stone ledge, the artist connects the picture’s space to our own, making the message all the more compelling.}
    \label{fig:ClaeszStillLifeSkullQuill}
\end{figure}

\begin{figure}
    \centering
    \includegraphics[scale=1.4]{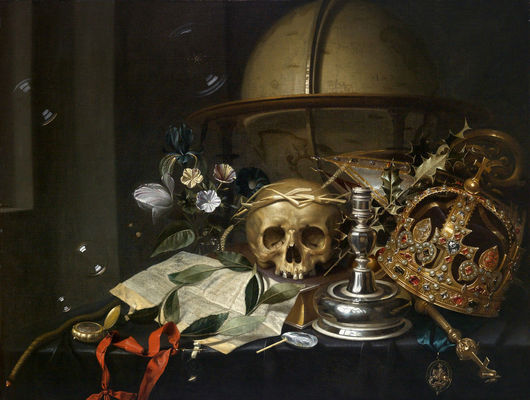}
    \caption{\textit{Vanitas Still-Life
Hendrick Andriessen, ca 1650} Before you read further, take a few moments to just look at this remarkably detailed painting. As your eyes move across the canvas, try to identify the objects that you see on the table. Often when we think of a still-life, we imagine a painting of fruit or flowers, so you might be surprised by some of the objects you see. This is a vanitas—a specific type of still-life that emerged in the 17th century in the Netherlands and grew out of a long artistic tradition known as memento mori, meaning “reminders of mortality.” While looking closely at this painting, you probably noticed several objects that could be called reminders of mortality, such as the skull, the wilting tulip, and the dying wick of the candle. \\Vanitas still-lifes were appreciated for both their beauty, rendered in incredible detail, and for their deeper symbolic significance. Andriessen’s contemporary audience may have recognized the crown as a specific, haunting reference to the recent execution of King Charles I of England in 1649. Every element of this painting also has broad symbolic power: the skull, bubbles, extinguished candle, flowers, and glass vase remind the viewer of the impermanence of life; the watch symbolizes the passing of time; the jeweled crown and bishop’s mitre lying behind it point to the fleeting nature of power.}
    \label{fig:AndriessenVanitasStillLife}
\end{figure}

\begin{figure}
    \centering
    \includegraphics[scale=0.5]{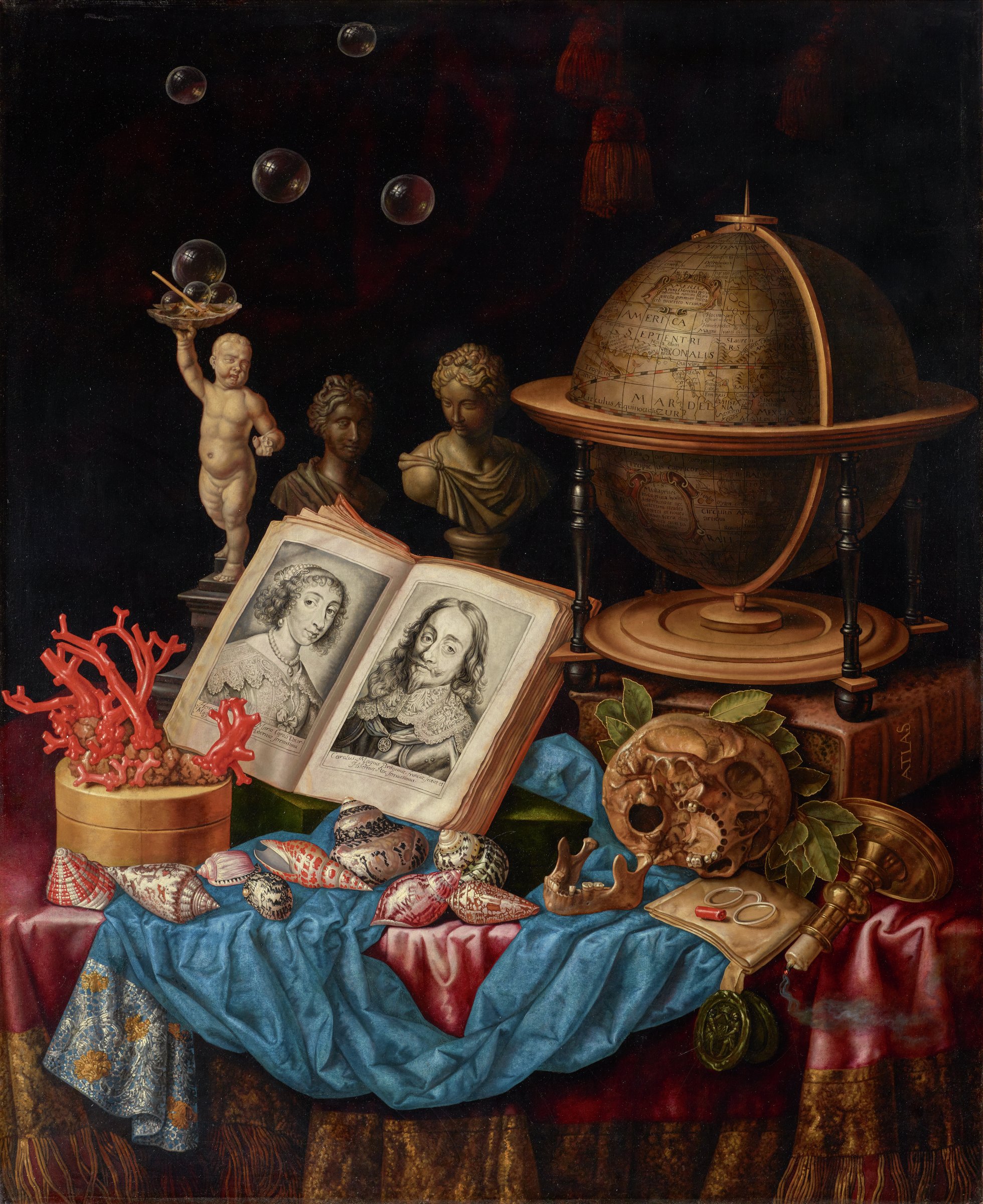}
    \caption{\textit{Vanitas Still Life
Unknown artist, possibly Flemish
After 1649} The subject here refers to Charles I’s troubled life and is a testament to the suddenness of death and the vanity of early power and glory. For example, the bubbles pertain to the brevity of Charles I’s life (he was beheaded at the age of 44), the broken skull conveys the fragility of human beings, and the globe symbolizes the power and possessions that death steals away.}
    \label{fig:UnknownVanitasStillLife}
\end{figure}

\begin{figure}
    \centering
    \includegraphics[scale=1.0]{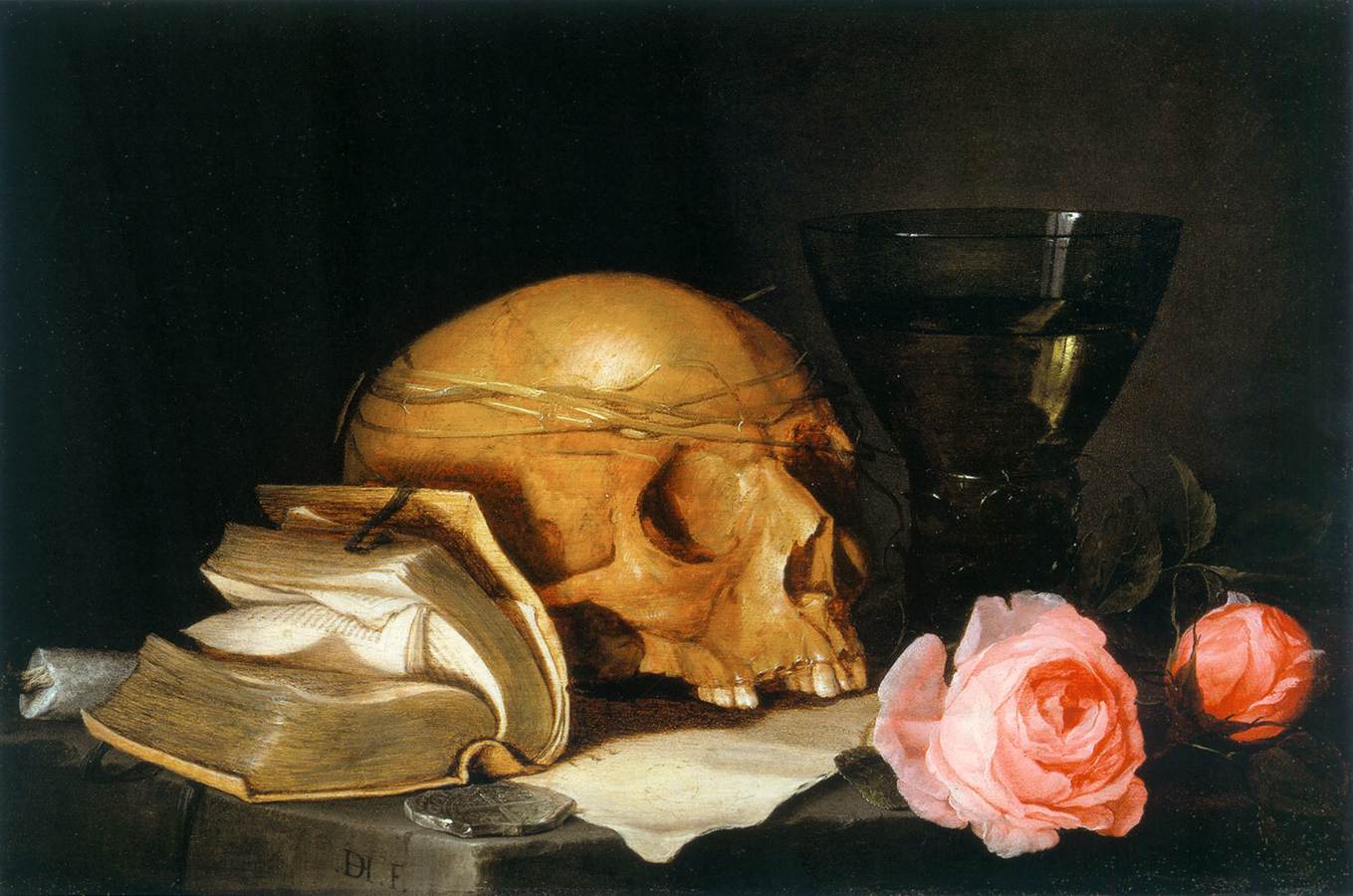}
    \caption{\textit{Heem, Jan Davidsz. de, Still-Life, c. 1630} This picture, showing a skull, a book and roses, is a memento mori, presenting a sinister contrast between the skull with its empty sockets and the fragrant pink rose so full of life.}
    \label{fig:HeemStillLife}
\end{figure}

\begin{figure}
    \centering
    \includegraphics[scale=1.0]{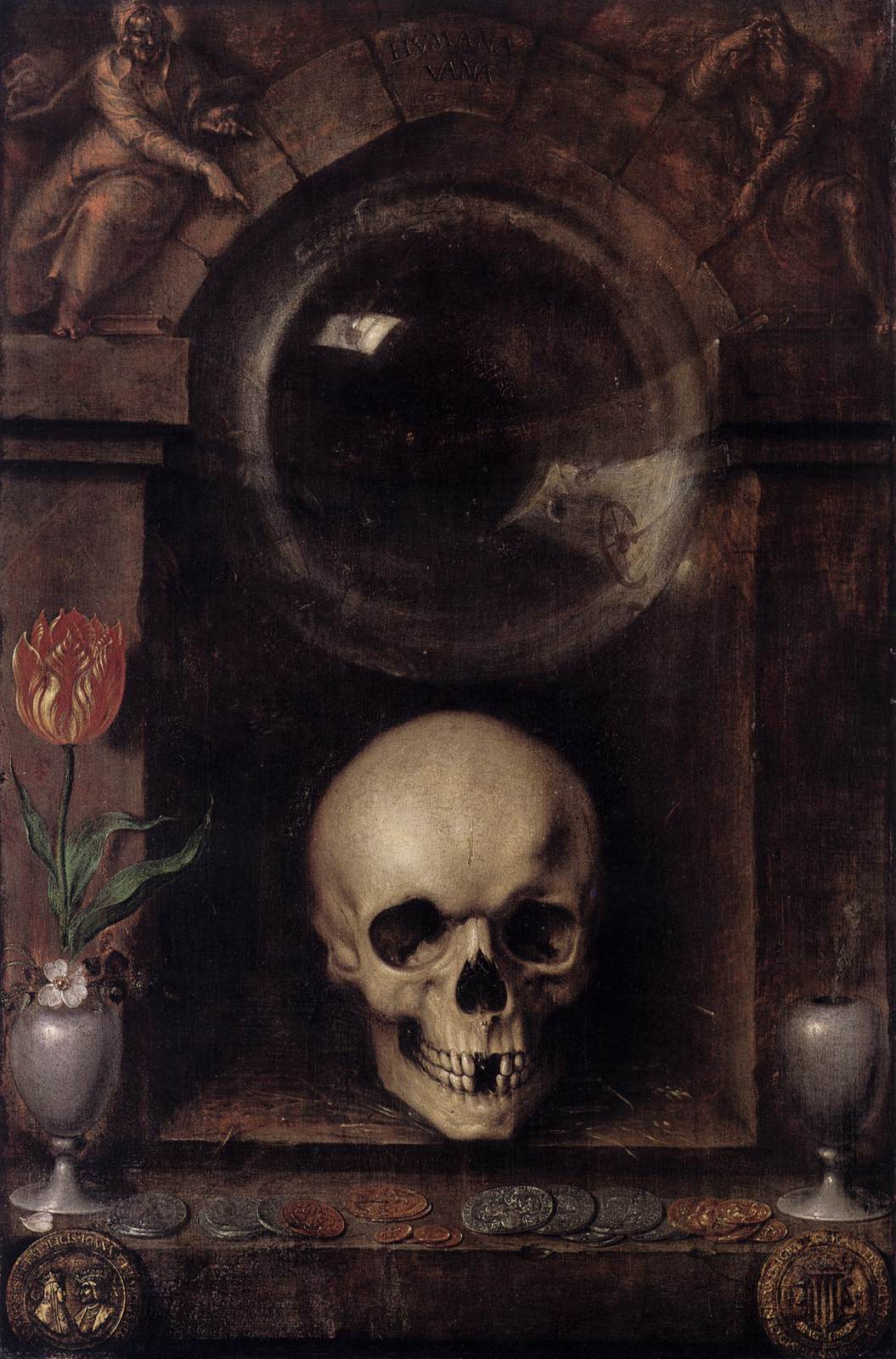}
    \caption{\textit{Gheyn, Jacob de II, Vanitas Still-Life, 1603} The dominant motifs in the picture are a human skull and, floating above it, a transparent sphere or bubble. These forms occupy a stone niche with a slightly pointed arch, the keystone of which is inscribed HVMANA VANA (Human Vanity). The spandrels flanking the arch are filled with sculptural figures of philosophers with books at their feet? to the left, Democritus, who gestures toward the globe and laughs; and, to the right, Heraclitus, who points to the sphere and weeps. The sphere purposefully resembles a soap bubble, the familiar vanitas motif. Two common vanitas symbols, cut flowers and smoke, rise from urns at either side of the niche. The coins depicted at the bottom of the composition on the sill between the vases were used as currency in the Netherlands about 1600. One of them, the silver medal of 1602 commemorates the capture of a Portuguese galleon by two Zeeland merchant ships earlier that year, off Saint Helena in the South Atlantic.}
    \label{fig:GheynVanitasStillLife}
\end{figure}

\begin{quote}
    These objects symbolize transitory human achievement and satisfactions. The atlas is open to a map of the East Indies, source of many Dutch fortunes, and there is a city council document with an imposing seal. The other open book is a history of the early counts of Holland-whose lands were absorbed by the dukes of Burgundy in the 1400s. The lute, music, and inkstand represent creative endeavors, which, like satisfaction in beautiful objects such as pearls, are transitory pleasures. Even the heavens and the earth, represented by two globes, are effected by Time, whose relentless passage is marked by the hourglass. 
\end{quote}

\section{System architecture} \label{sec:Architecture}

\begin{figure}
    \centering
    \includegraphics[scale=0.3]{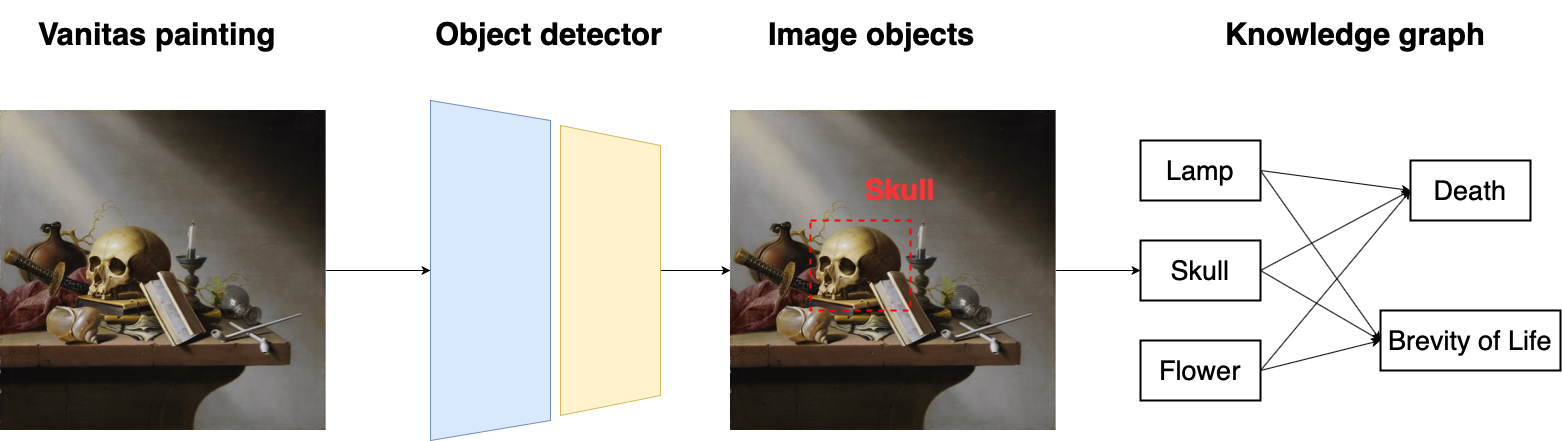}
    \caption{System architecture.  The painting is provided as input, which is then processed to recognize and locate the objects.  Each recognized object is then passed to the knowledge graph, which links that object (signifier) to its meaning (signified).}
    \label{fig:SystemArchitecture}
\end{figure}

Our system architecture, shown in Fig.~\ref{fig:SystemArchitecture}, consists of image analysis and natural language processing.

Figure~\ref{fig:KnowledgeGraph} shows an example of a knowledge graph that could be produced using the natural language processing (NLP) techniques described below.


\subsection{Visual processing} \label{sec:VisualProcessing}

A key component of our system is a deep convolutional network object identification and localization module. We take an implementation of mask R-CNN \cite{2017_Mask, 2017_repo} that has been pre-trained on the Microsoft COCO (Common Objects in Context) dataset \cite{2014_Lin} and apply transfer learning to our dataset of Vanitas images.

\subsection{Natural language processing} \label{sec:NaturalLanguage}

\begin{figure}
    \centering
    \includegraphics[scale=0.4]{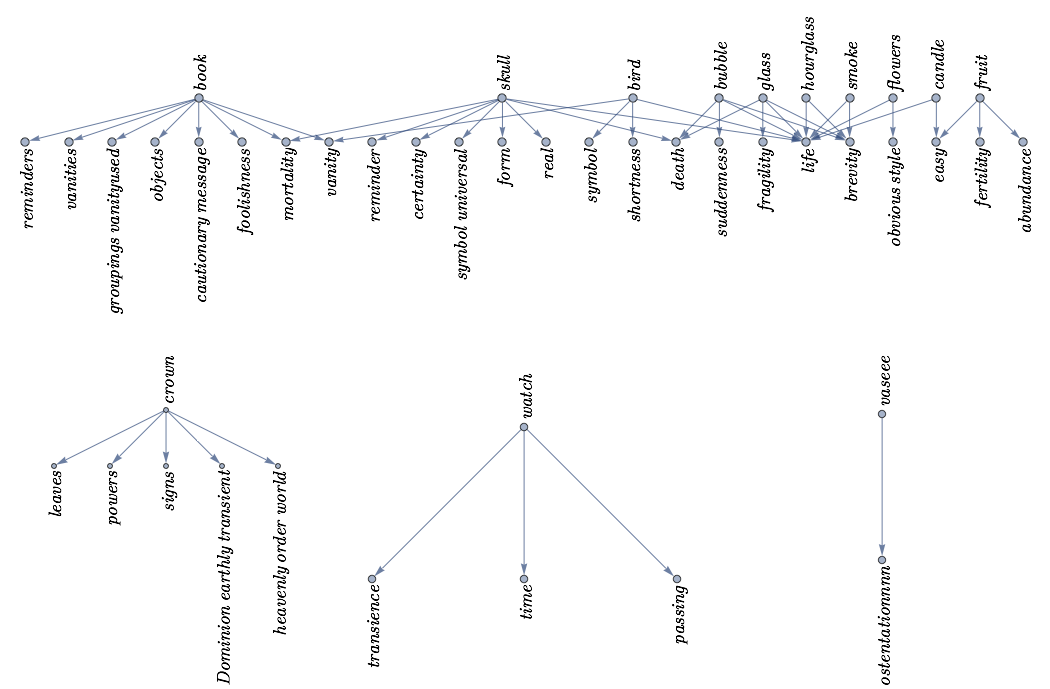}
    \caption{Our bipartite knowledge graph linking signifiers to signifieds, learned from textual descriptions of vanitas paintings written by contemporaneous critics and art historians.}
    \label{fig:KnowledgeGraph}
\end{figure}

The objects detected by the vision component can be mapped to abstract meanings using a knowledge graph.  An example of one that was generated automatically is shown in Fig.~\ref{fig:KnowledgeGraph}.

\section{Training and testing} \label{sec:TrainAndTest}


\subsection{Natural language processing}

The knowledge graph was created using texts from pages on the internet.  This involved retrieving the URLs of the websites from Google using the query \lq\lq vanitas meaning\rq\rq\ and scraping the corresponding websites.  The total number of pages scraped was 96.

The texts containing references to the artists and paintings included in the test set for the vision modules was reserved for evaluating the knowledge graph (27 pages).  The rest of the texts were used as the train set for constructing the graph (69 pages).

\subsection{Training} \label{sec:Training}

\subsection{Natural language processing}

The object-meaning pairings were extracted by performing semantic role labelling on the texts using a BERT-based model \cite{DBLP:journals/corr/abs-1904-05255}.  The heads (objects) of the knowledge graph were set to be the proto-agents (Arg0) or if none were available, the proto-patients (Arg1).  All the other arguments were assigned as tails (meanings) of each corresponding head (given a predicate).  Each edge was weighted by the number of times a head-tail paring appeared in the texts, i.e., an appearance of 2 corresponded to a weight of 2.  The knowledge graph was pruned by including only the heads that represented the objects detected by the vision module.  The tails that corresponded to undesirable entities, i.e., person, organisation, geopolitical entity, location, were excluded. The transition-based entity recognizer provided by the Spacy library was used for this purpose.  Finally, any edges with a weight less than 2 were removed.

\subsection{Visual processing}

We transfer learn the weights of the pre-trained mask R-CNN model on 56 images of vanitas paintings. The training procedure operates in two phases: First, we freeze the weights of all layers save for the final layer of mask R-CNN and fine-tune the weights of this final layer for 10 epochs using a learning rate of $0.001$. Second, we fine-tune all layers of the network for 2 epochs using a learning rate of $0.0001$. Training is performed on a single NVIDIA Tesla K80 GPU in the Google Colab environment.

\subsection{Testing} \label{sec:Testing}

\subsection{Natural language processing}

The 27 texts in the test set were used to extract the object-meanings pairings manually.  This led to the creation of a subset that matched the objects and meanings from descriptions in specific paintings and of a another consisting of more general pairings (i.e., what associations between objects and meanings existed for vanitas paintings in general).  
When evaluating the knowledge graph exact matches and partial matches between the meanings of the knowledge graph and the test set were used to calculate the precision, recall and F1 score.  An exact match was defined as the instance where a meaning in the knowledge graph (i.e., \lq life\rq ) and the meaning in the test set (i.e., \lq life\rq ) were identical.  A partial match was when a meaning in the knowledge graph (i.e., \lq life\rq ) was a subset of the meaning in the test set (i.e., \lq brevity of life\rq ).

Additionally, the semantic similarity between the meanings in the knowledge and the test set was used to define matches (true positives), i.e., if the cosine distance between the BERT sentence embeddings \cite{DBLP:journals/corr/abs-1908-10084} was above a predefined threshold ($0.7$) (i.e. \lq time\rq\ and \lq transience\rq ).  If the similarity between a meaning in the knowledge and all of the meanings in the test set for a given object was below the threshold, the meaning in the knowledge graph was deemed a false positive.  Likewise, if the similarity between a meaning in the test and all the meanings in the knowledge for a given object was below the threshold, the meaning in the test set was deemed a false negative.

\subsection{Visual processing}

We test the mask R-CNN object detector module on 18 heldout images of Vanitas paintings. As with training, inference is again performed on a single NVIDIA Tesla K80 GPU in the Google Colab environment. All code to reproduce our results is made available at \href{https://github.com/gck25/fine_art_asssociations_meanings}{https://github.com/gck25/fine\_art\_asssociations\_meanings}.

\subsection{End-to-end system}

The full system was tested on 8 images from the test set for which descriptions relating to each specific image could be found in the web texts.  The system\rq s ability to map paintings to meanings was assessed.  The same metrics used to evaluate the NLP component was used to evaluate the end-to-end system.

\section{Results} \label{sec:Results}

We express our results using {\em precision} and {\em recall}, where 

\begin{equation}
    precision = \frac{TP}{TP + FP},\quad\quad  recall = \frac{TP}{TP + FN}
\end{equation}

\noindent and $TP$ are true positives, $FP$ are false positives, and $FN$ are false negatives.

\subsection{Natural language module and processing}

The semantic and partial matches shown in Table~\ref{table:results} are significantly higher than the exact matches.  This suggests that although the knowledge graph does not associate objects with exact phrases in the test set, it does connect them to meanings that are semantically similar.

\subsection{Vision module and processing}

According to Table~\ref{table:results}, the precision is substantially larger than the recall.  The majority of the tags assigned by the component are correct, but it is missing tags for a large number of objects.

\subsection{End-to-end system}

The end-to-end system follows the same same trend as the knowledge graph in terms of the relative difference in the metrics between exact, partial and semantic matches.  Nonetheless, the recall is larger than the precision.  This could be explained by the difference in test sets used to evaluate each component.  The knowledge graph test set was created using texts that described Vanitas symbolism in general, while the end-to-end test set used texts referring to specific images.  The painting-specific texts associate a lower number of meanings to each image, which increase the recall and lower the precision.

\begin{table}
\caption{Precision, recall, and F1 performance for each component in the overall system.}
\label{table:results}
\centering
\begin{tabular}{|c c c c|} 
 \hline
 Component & Precision & Recall & F1 \\ [0.5ex] 
 \hline\hline
 Knowledge Graph (exact) & 0.11 & 0.04 & 0.06 \\
  \hline
 Knowledge Graph (partial) & 0.26 & 0.18 & 0.22  \\
 \hline
 Knowledge Graph (semantic) & 0.49 & 0.39 & 0.43  \\ 
 \hline
 Object Detector & 0.6 & 0.06 & 0.10 \\
  \hline
 End-to-end System (exact) & 0.04 & 0.11 & 0.06 \\
 \hline
 End-to-end System (partial) & 0.12 & 0.33 & 0.17 \\
 \hline
 End-to-end System (semantic) & 0.48 & 0.78 & 0.60 \\
 \hline
%
\end{tabular}
\end{table}

\section{Conclusions and future directions} \label{sec:Conclusions}

We have demonstrated computational approaches to the extraction of simple meanings in authored images, specifically vanitas paintings from the Dutch Golden Age.  Our method involves natural language processing of contemporaneous and scholarly texts commenting on such artworks, to build a bipartite knowledge graph relating visual signs in the paintings to signifiers.  Our end-to-end system had precision and recall of $0.48$ and $0.78$, respectively.  Our work may serve as a base for higher-level semantic interpretations of images, including additional genres, such as religious art, mythological art, print advertisements, and political propaganda.

Our work is a very early---yet, we feel, promising---step in a broad program of understanding author-created meanings in images.  Art is a rich source of problems in this domain and far richer than the domain of natural photographs, which dominates research in computer vision.  This nascent research program expands the class of problems faced by AI and, once enhanced, may serve as a tool for art scholars.\cite{Stork:22} 

\section*{Acknowlegements}

The last author would like to thank the Getty Research Institute, Los Angeles, where some of the work on this project was performed.

\section*{Code available}

Our code is available at:  \url{https://github.com/gck25/fine_art_asssociations_meanings}.

\bibliography{Art}
\bibliographystyle{spiebib}

\end{document}